\documentclass[conference]{IEEEtran}
\IEEEoverridecommandlockouts
\usepackage{cite}
\usepackage{amsmath,amssymb,amsfonts}
\usepackage{algorithmic}
\usepackage{graphicx}
\usepackage{subcaption}
\usepackage{textcomp}
\usepackage{xcolor}
\usepackage{siunitx}
\usepackage{mathtools}
\usepackage{esint}
\usepackage{hyperref}
\usepackage{multirow}
\def\BibTeX{{\rm B\kern-.05em{\sc i\kern-.025em b}\kern-.08em
    T\kern-.1667em\lower.7ex\hbox{E}\kern-.125emX}}

\newcommand*\Laplace{\mathop{}\!\mathbin\bigtriangleup}

\newcommand{\Unit}[1]{\thinspace #1}

\begin{document}

\title{Windscreen Optical Quality for AI Algorithms: Refractive Power and MTF not Sufficient} 

\author{\IEEEauthorblockN{1\textsuperscript{st} Dominik Werner Wolf}
\IEEEauthorblockA{\textit{Glass Laboratory} \\
\textit{Volkswagen Group}\\
Wolfsburg, Germany \\
dominik.werner.wolf@volkswagen.de}
\and
\IEEEauthorblockN{2\textsuperscript{nd} Markus Ulrich}
\IEEEauthorblockA{\textit{Inst. of Photogram. and Remote Sens.} \\
\textit{Karlsruhe Institute of Technology}\\
Karlsruhe, Germany \\
markus.ulrich@kit.edu}
\and
\IEEEauthorblockN{3\textsuperscript{rd} Alexander Braun}
\IEEEauthorblockA{\textit{Faculty of Electr. Eng. and Info. Tech.} \\
\textit{University of Applied Sciences Duesseldorf}\\
Duesseldorf, Germany \\
alexander.braun@hs-duesseldorf.de}
}

\maketitle

\begin{abstract}
Windscreen optical quality is an important aspect of any advanced driver assistance system, and also for future autonomous driving, as today at least some cameras of the sensor suite are situated behind the windscreen. Automotive mass production processes require measurement systems that characterize the optical quality of the windscreens in a meaningful way, which for modern perception stacks implies meaningful for artificial intelligence (AI) algorithms. The measured optical quality needs to be linked to the performance of these algorithms, such that performance limits -- and thus production tolerance limits -- can be defined. In this article we demonstrate that the main metric established in the industry -- refractive power -- is fundamentally not capable of capturing relevant optical properties of windscreens. Further, as the industry is moving towards the modulation transfer function (MTF) as an alternative, we mathematically show that this metric cannot be used on windscreens alone, but that the windscreen forms a novel optical system together with the optics of the camera system. Hence, the required goal of a qualification system that is installed at the windscreen supplier and independently measures the optical quality cannot be achieved using MTF. We propose a novel concept to determine the optical quality of windscreens and to use simulation to link this optical quality to the performance of AI algorithms, which can hopefully lead to novel inspection systems.
\end{abstract}


\section{Introduction}
Every car has a windscreen. The number of newly produced windscreens therefore ranges in the millions every year. Following quality processes for automotive mass production established since the 1960ies -- like the outdated ISO/TS 16949 \cite{ISO16949} or the more recent VDA6.3 \cite{VDA6.3} -- these windscreens are tested end-of-line (EOL) at the suppliers (Tier~1) production line using well-defined optical measurements. Importantly, the windscreen quality is measured at the production site alone, independent of any production tolerances that may arise during assembly of the whole car at the site of the car manufacturer (original equipment manufacturer, OEM). Economically, this is mandatory, as a thorough testing of the whole windscreen after assembly by the OEM is prohibitively expensive.

For several decades the optical quality of these windscreens has been judged acceptable if humans could look through it with low impact on the perception of the driver. With the rise of advanced driver assistance systems (ADAS) and the future promise of autonomous driving (AD) many cars nowadays are equipped with several camera systems, many of which are situated behind the windscreen. A camera is not a human observer, and it is now not enough to qualify a windscreen using human perception, especially as the quality and resolution of the cameras are steadily increasing. The influence of the optical quality on the image quality and further on the computer vision algorithms evaluating these images has to be precisely determined.

In theory, the working limits of the computer vision algorithms are determined, and production tolerance limits are derived from these algorithmic working limits through a number of processes defined in the above mentioned quality norms. Opto-mechanical tolerance calculations, numerical simulations and test campaigns in the real world form three important pillars of these studies accompanied by environmental stress tests and aging simulations \cite{
Windscreens_and_safety_II, Windscreens_and_safety_III}.

In practice, though, modern camera-based ADAS applications are based on artificial intelligence (AI) and employ deep convolutional neural networks, due to their superior performance in comparison to traditional, rule-based computer vision algorithms. The difference in performance is such that currently there is no alternative to using AI algorithms. As these AI algorithms are 'black boxes' in nature \cite{heizmann:22}, i.e. the output cannot be predicted, the link between optical quality and AI algorithm performance cannot be easily established \cite{Braun_MTF_performance}. And due to the lack of quantitative working limits w.r.t. the AI algorithms, production tolerance limits for the windscreens can not be straightforwardly deduced \cite{Braun_tm_2022}.

In this article we are evaluating the two main measurement processes that are currently used in the automotive industry to qualify windscreen optical quality: refractive power and the modulation transfer function (MTF). While refractive power is the established measurement method and has been standardized already in the 1990ies \cite{DIN52305, ECE43}, the MTF -- or equivalently the spatial frequency response (SFR) -- has gained recent attention as automotive researchers \cite{BMW} look for alternatives to refractive power because of the increasing ADAS camera performances in terms of the number of pixels per field angle. Novel startups are even forming around the promise of using MTF to characterize windscreens.

We find and mathematically demonstrate in this work that both refractive power and MTF are not sufficient to quantify windscreen quality for AI algorithm performance. This is a fundamental finding in that our results are derived from first principles of optics, and apply very generally. First, we recapitulate the optical basics in Sec.~\ref{sec:opticsBasics}. Importantly, the optical quality is described in terms of wavefront aberrations, using the Zernike formalism to mathematically decompose the nature of the optical perturbations. Then in Sec.~\ref{sec:refractivePower}, using these basics we show how refractive power is fundamentally not capable of accounting for a distinct number of wavefront aberrations, while at the same time these aberrations have a demonstrable effect on AI algorithm performance \cite{Mueller_dist_shift, Mueller_sim}.

In Sec.~\ref{sec:MTF} we then show how the windscreen and the camera system form a joint optical system, that -- again fundamentally -- cannot be separated into two distinct optical systems. This separation, though, is a necessary requirement in linear system theory for the multiplicativity of the system MTF w.r.t. the individual optical elements \cite{Eugene_Hecht}. Therefore, this prohibits any MTF measurement on the windscreen alone, and thus from using MTF as a qualifying measurement at the production site of the Tier~1.

Optical quality has many different aspects. For this article, we concentrate solely on the 'sharpness' of the camera image, which is deteriorated by optical path variations across the windshield plane and is typically quantified by the MTF in optical linear system theory. In general, lens distortions, which describe the failure of a lens to map lines into lines and represent a curvilinear mapping \cite{steger:18a}, might also deteriorate the performance of ADAS functionalities. Effects of optical distortions will not be considered in the following.


In summary, we will show how the two only current measurement techniques in the automotive industry are not sufficient to measure the sharpness of the windscreen alone. These results have far reaching implications for the automotive industry, which needs to focus more effort on finding alternatives. We finally propose a concept on how to find a novel measurement process, combining optical modeling, numerical simulation and AI algorithms to link the optical quality of windscreens to the performance of AI algorithms.


\section{Optical Quality and Mathematical Models}
\label{sec:opticsBasics}

Maxwell's equations are the fundamental physical model of electromagnetic radiation, and the wave equation forms the basis for the technological application of light. If all elements in the optical system are large compared to the wavelength of the light, geometrical optics may be used. It plays an important role in the development of optical systems as well, in the form of raytracing simulations. A windscreen is large in mechanical dimensions, both laterally as well as axially, but previous work has shown that the aberrations originating inside the windscreen cannot be neglected \cite{krebs, Lumetrics}. Thus, it is not sufficient to take only the geometry of the windscreen into account -- which would allow for a raytracing approach -- but a comprehensive optical model needs to be based on the wave description of light. This is why in the following we use the fundamental Zernike approach \cite{Goodman} to model wavefront aberrations, where the optical path difference mathematically models the aberrations present in the windscreen.

\subsection{Wavefront Modelling with Zernike Polynomials}
The optical path difference $W$, defined on the principle plane, is usually expressed as a decomposition into Zernike polynomials $Z_{n}$ with corresponding Zernike coefficients $c_{n}$ (in units of meters) as~\cite{born_wolf}:
\noindent
\begin{equation}
    W(\rho,\;\phi) = \sum \limits_{n=0}^{\infty} c_{n} Z_{n}(\rho,\;\phi) \;\;\;\text{,}\;\;\; c_{n} :\overset{(\ref{eq:Zernike_scalar_product})}{=} \left<W,\;Z_{n}\right>\;.
\label{eq:Zernike_decomposition}
\end{equation}
\noindent
Here, the domain of the principle plane of the optical element is parameterized by normalized polar coordinates with radius $\rho$ and polar angle $\phi$. There are different numbering schemes for Zernike polynomials, i.a. a linear numbering scheme according to the American National Standards Institute (ANSI) which has been adopted within this work. The Zernike polynomials reproduce the aberration pattern on the unit circle and correspond to different, independent optical perturbations like defocus or astigmatism.

The independence of the perturbations is mathematically reflected by the orthogonality relation of the scalar product:
\noindent
\begin{equation}
    \left< Z_{i} ,\; Z_{j} \right> \coloneqq \int \limits_{0}^{2\pi} \int \limits_{0}^{1} Z_{i}(\rho,\;\phi) \cdot Z_{j}(\rho,\;\phi) \cdot \rho \; \mathrm{d}\rho \; \mathrm{d}\phi = \pi \cdot \delta_{ij}\;.
\label{eq:Zernike_scalar_product}
\end{equation}
\noindent
This is important, because we will demonstrate how certain Zernike polynomials are simply not present in the refractive power measurement, and the orthogonality fundamentally implies that this information can not be recovered. Table~\ref{tab:Zernike_polynomials} indicates the normalized Zernike polynomials defined by ISO 24157 \cite{ISO24157} up to the third order.
\noindent
\begin{table}[!b]
    \centering
    \resizebox{\linewidth}{!}{
    \begin{tabular}{c|l|l|c}
        \hline
        $Z_{i}$ & \multicolumn{2}{c}{Zernike polynomial} & Harmonic\\
        & Polar coordinates & Cartesian coordinates &\\
        \hline
        \hline
        $Z_{0}$ & $1$ & $1$ & \checkmark \\
        $Z_{1}$ & $2\rho \sin\phi$ & $2y$ & \checkmark \\
        $Z_{2}$ & $2\rho \cos\phi$ & $2x$ & \checkmark \\
        $Z_{3}$ & $\sqrt{6}\rho^2 \sin2\phi$ & $2\sqrt{6}xy$ & \checkmark \\
        $Z_{4}$ & $\sqrt{3}(2\rho^2 - 1)$ & $\sqrt{3}(2x^2 + 2y^2 - 1)$ & $\mathbf{\times}$ \\
        $Z_{5}$ & $\sqrt{6}\rho^2 \cos2\phi$ & $\sqrt{6}(x^2 - y^2)$ & \checkmark \\
        $Z_{6}$ & $\sqrt{8}\rho^3 \sin3\phi$ & $\sqrt{8}(3x^2y - y^3)$ & \checkmark \\
        $Z_{7}$ & $\sqrt{8}(3\rho^3 - 2\rho)\sin\phi$ & $\sqrt{8}(3x^2y + 3y^3 - 2y)$ & $\mathbf{\times}$ \\
        $Z_{8}$ & $\sqrt{8}(3\rho^3 - 2\rho)\cos\phi$ & $\sqrt{8}(3x^3 + 3xy^2 - 2x)$ & $\mathbf{\times}$ \\
        $Z_{9}$ & $\sqrt{8}\rho^3 \cos3\phi$ & $\sqrt{8}(x^3 + 3xy^2)$ & \checkmark \\
        \hline
    \end{tabular}}
    \vspace{2mm}
    \caption{Zernike polynomials up to the third order.}
    \label{tab:Zernike_polynomials}
\end{table}

\subsection{Refractive Power}
Refractive power measures how much focusing power a lens has. It is given in units of diopters, i.e. in inverse distance of the focal length of the lens. A comprehensible way to visualize refractive power is two parallel light rays entering the optical system -- here: the windscreen -- and upon exit are not parallel anymore, but either divergent or convergent. In the convergent case, the focal length is the distance from the refractive element to the intersection of the two rays, and its inverse is the numerical value of the refractive power. For concave lenses, the diverging rays are extended in the negative direction until these two rays intersect, and the negative distance now forms the focal length.
\begin{figure}[!t]
  \centering
   \includegraphics[width=1.\linewidth]{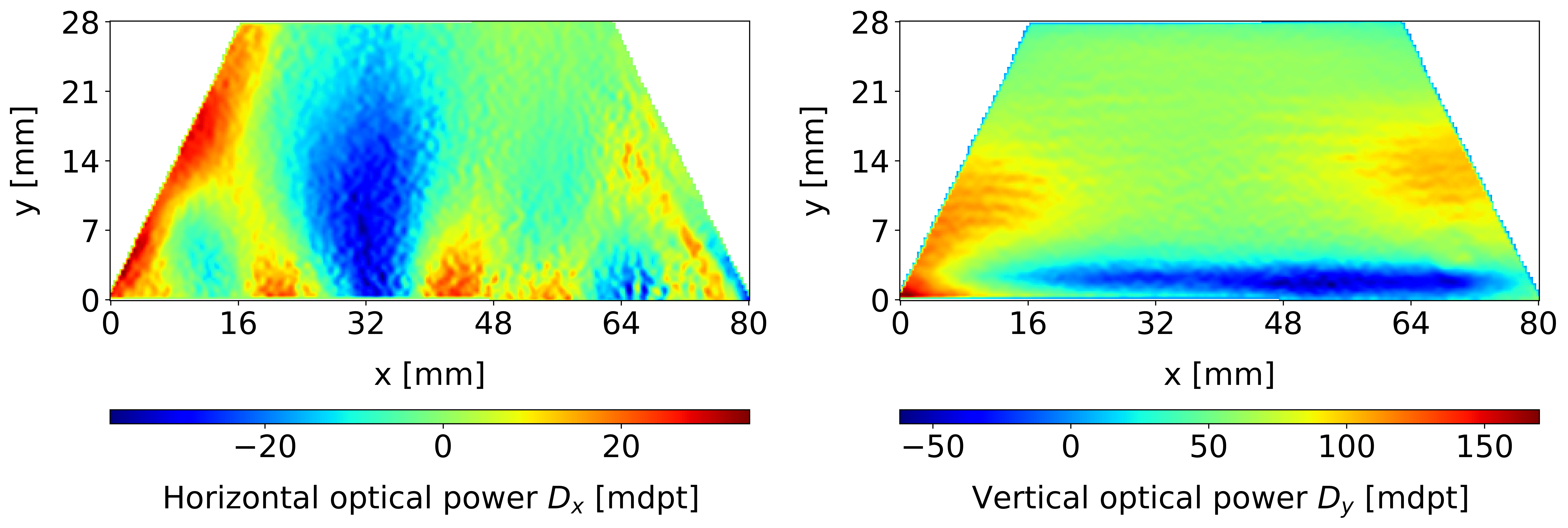}
   \caption{Refractive power measurement of the ADAS camera area of a VW series production windshield under an inclination angle of $\epsilon = 63$°. The difference in magnitude between horizontal and vertical direction originates from the inclination angle, which amplifies the refractive power according to the Kerkhof model \cite{Metrologia}.}
   \label{fig:optical_power_measurement}
   \vspace{-0.5cm}
\end{figure}

For windscreens, the refractive power is not a single number for the whole glass, but the measurement has a spatial resolution as depicted by Fig.~\ref{fig:optical_power_measurement}. In the early days, two actual parallel laser beams were deflected, and the whole setup was laterally moved to achieve a certain spatial resolution \cite{DIN52305}. More modern systems such as the one produced by ISRA use the Moiré effect to spatially resolve the refractive power over a limited area by observing the location dependency of the perturbed grid spacing between Moiré interferences \cite{ISRA}. In addition, new refractive power measurement systems like the one produced by LaVision \cite{LaVision} use the Background Oriented Schlieren (BOS) imaging method to overcome the resolution limitation of the Moiré approach \cite{Metrologia}.

Importantly, the refractive power depends on the direction, as the two parallel lines form a plane together with the principal plane of the optical element. In principle, this direction can be rotated full circle by \SI{360}{\degree}, but in practice, the refractive power is determined and specified only in the horizontal and vertical direction.

\subsection{Modulation Transfer Function}
\label{sec:MTF_intro}
The modulation transfer function (MTF) -- and its non-harmonic equivalent, the spatial frequency response (SFR) -- are established metrics to characterize optical systems, based on linear system theory and scalar diffraction theory \cite{Boreman, Goodman}. In image space, the transfer function of the system under test is called the point spread function (PSF), and in frequency space, it is denoted as the optical transfer function (OTF). The MTF is given by the absolute value of the OTF and is of particular importance if the intensity distribution is the matter of interest. The PSF and the MTF are highly non-linear functions over the image field (radius, azimuth), and they also depend on the defocus $\Delta z$, due to the refractions on different lens element surfaces. Hence, the input space of the PSF is in general three dimensional.

The MTF is measured by using either harmonic input signals (MTF, e.g. sinusoidal Siemens star) or a step function type input (SFR, e.g. slanted edge). ISO12233 defines a norm to measure the MTF \cite{ISO12233}, and IEEE P2020 is currently finalizing an automotive extension of this norm \cite{IEEEP2020}. In this article, we will use slanted edge measurements.

According to scalar diffraction theory, the MTF is proportional to the absolute value of the Fourier transform of the wavefront in the aperture plane of the lens (more general: the optical element). The wavefront is transformed, normalized, and the absolute value is taken to yield the MTF. This allows for an analytical relationship between the MTF and the wavefront aberrations, which can be parameterized by Zernike coefficients $c_{n}$:
\noindent
\begin{equation}
     \resizebox{0.9\linewidth}{!}{$\mathrm{MTF} (\vec{k}|\lambda) = \left| \dfrac{\resizebox{0.15\linewidth}{!}{$\oiint\limits_{P_{+}\cap P_{-}}$} \exp{\left(\dfrac{2\pi\mathrm{i}}{\lambda} \sum \limits_{n=0}^{\infty} c_{n} \left[ Z_{n}(\vec{\xi} + \vec{\Delta}) - Z_{n}(\vec{\xi} - \vec{\Delta}) \right]\right)} \; \mathrm{d\xi^2}}{\resizebox{0.1\linewidth}{!}{$\iint_{\mathbb{R}^2}$} |\;P(\vec{\xi})\;|^2 \; \mathrm{d\xi^2}} \right|\;.$}
\label{eq:Zernike_parameterization}
\end{equation}
\noindent
Eq.~(\ref{eq:Zernike_parameterization}) is motivated by Goodman \cite{Goodman} and the domain of integration is determined by the aperture stop of the camera. In detail, $P$ describes the 2D aperture stop function, which is given by a circular top hat function with magnitude one and baseline zero. The displaced aperture stop function $P_{+}$ is shifted by:
\noindent
\begin{equation}
    \Vec{\Delta} \coloneqq \lambda z_{a \mapsto o} \dfrac{\Vec{k}}{2} \approx \lambda f \dfrac{\Vec{k}}{2}\;.
\label{eq:displaced_aperture_stop_function}
\end{equation}
\noindent
The intersection of $P_{+}$ and $P_{-}$ (shift by $-\vec{\Delta}$) determines the domain of integration in Eq.~(\ref{eq:Zernike_parameterization}). In addition, $\lambda$ characterizes the wavelength under consideration and $z_{a \mapsto o}$ quantifies the distance on the optical axis from the aperture stop to the observer plane, which roughly equals the focal length $f$ of the camera lens if the Gaussian lens equation is approximated. This simplification holds if the working distance is by magnitudes larger than the image distance $z_{a \mapsto o}$. Finally, $\Vec{k}$ denotes the spatial frequency vector.

As a side note, the $\mathrm{MTF}$ depends on the wavelength $\lambda$ according to Eq.~(\ref{eq:Zernike_parameterization}). The polychromatic $\mathrm{MTF}$ can be retrieved by integrating the conditioned, monochromatic $\mathrm{MTF}$ over the normalized power spectral density $\mathrm{PSD}$ of the light source. The area under the $\mathrm{PSD}$ curve quantifies the likelihood of emitting a photon in the wavelength range $[\lambda,\;\lambda + \Delta \lambda]$ by the light source. As a result, the polychromatic $\mathrm{MTF}$ is given by:
\noindent
\begin{equation}
    \mathrm{MTF} (\vec{k}) = \int \limits_{0}^{\infty} \mathrm{MTF}(\vec{k}|\lambda) \cdot \mathrm{PSD}(\lambda) \; \mathrm{d}\lambda\;.
\label{eq:poly_MTF}
\end{equation}
\noindent
Consequently, chromatic aberrations will potentially also influence the performance of AI-based algorithms for autonomous driving but they are not discussed in detail in this paper.

At this point, it has to be emphasized, that the first three Zernike coefficients for piston and tilt do not represent optical aberrations in the classical sense. Even though there is a wavefront perturbation of the light beam, the image quality is not influenced by those terms because the curvature of the wavefront is not affected. Instead, tilts induce image distortions and generate a non-conformal mapping. As we focus in this article on the sharpness of the optical system we will not further investigate this influence.

If the Zernike polynomials of Table~\ref{tab:Zernike_polynomials} are transformed into Cartesian coordinates it becomes obvious that the difference in Eq.~(\ref{eq:Zernike_parameterization}) vanishes for Zernike polynomials of zeroth order. For the y-tilt $Z_{1}$ and the x-tilt $Z_{2}$, the integrand evaluates to a constant phasor. Hence, it holds that:
\noindent
\begin{align}
\begin{split}
     \mathrm{MTF} (\vec{k}|\lambda) &\overset{(\ref{eq:Zernike_parameterization})}{=} \left| \; \mathrm{e}^{\left(2\pi\mathrm{i} \cdot c_{1,2} \cdot z_{a \mapsto o} \cdot \Vec{k}\right)} \cdot \dfrac{\oiint\limits_{P_{+}\cap P_{-}} 1 \; \mathrm{d\xi^2}}{\iint_{\mathbb{R}^2} |\;P(\vec{\xi})\;|^2 \; \mathrm{d\xi^2}} \; \right|\\[2pt]
     \Leftrightarrow \mathrm{MTF} (\vec{k}|\lambda) &\;= \left| \; \mathrm{e}^{\left(2\pi\mathrm{i} \cdot c_{1,2} \cdot z_{a \mapsto o} \cdot \Vec{k}\right)} \; \right| \cdot \dfrac{\oiint\limits_{P_{+}\cap P_{-}} 1 \; \mathrm{d\xi^2}}{\iint_{\mathbb{R}^2} |\;P(\vec{\xi})\;|^2 \; \mathrm{d\xi^2}}\\[5pt] \Leftrightarrow \mathrm{MTF} (\vec{k}|\lambda) &\coloneqq \left| \; \mathrm{e}^{\left(2\pi\mathrm{i} \cdot c_{1,2} \cdot z_{a \mapsto o} \cdot \Vec{k}\right)} \; \right| \cdot \mathrm{MTF_{diff}} (\vec{k}|\lambda)\\[5pt]
     \Leftrightarrow \mathrm{MTF} (\vec{k}|\lambda) &\;= \mathrm{MTF_{diff}} (\vec{k}|\lambda)\;.
\end{split}
\label{eq:MTF_tilt}
\end{align}
\noindent
As a consequence, the diffraction limited $\mathrm{MTF}$ is not modulated by the Zernike coefficients $c_{0}$ up to $c_{2}$. Therefore, the second order Zernike coefficients are those of main interest for studying the effect of optical aberrations in terms of sharpness degradation on convolutional neural networks for autonomous driving. Nonetheless, the optical distortion does influence the rectilinear mapping, but as stated before this is an independent effect we do not study in this work.

Finally, note that the MTF is unfortunately currently not traceable to fundamental physical quantities, and therefore a calibration chain to national metrological institutes like the PTB in Germany or the NIST in the US can not be established at the moment. For the automotive industry, this is a major source of discussion, as the implementations of ISO12233 are very sensitive to many diverse influences: stable lighting conditions (spectrum, intensity, direction, homogeneity), a reproducible mechanical setup (target distance, field-of-view) and well-defined camera settings (gain, exposure, HDR, ISP \dots) are the goal, which are not met in practice. Instead, comparability of two camera systems is only possible to a (relatively) good accuracy within measurements from the very same experimental setup. Comparison between two different measurement sites -- even with the same nominal setup -- is quite difficult and error-prone \cite{MTF_wedges, MTF_stability_repeatability, MTF_aliasing}.

\section{Refractive Power}
\label{sec:refractivePower}
In this section, we will demonstrate how relevant Zernike polynomials are not captured by refractive power measurements. Our argument is based on the theory that the refractive power is given by the second derivative of the wavefront modulation of a plane wave passing a refractive element \cite{Wavefront_refractive_power}. Using this relationship, we demonstrate how several Zernike polynomials are simply not covered by a refractive power measurement, as zeroth and first order polynomials in $x$ and $y$ vanish if the second derivative is considered. We start by introducing the measurement principle of a Shack-Hartmann sensor in Sec.~\ref{sec:shackHartmann} to motivate how the Zernike coefficients $c_{i}$ are retrieved. In Sec.~\ref{sec:secondDerivative}, we then mathematically derive which Zernike polynomials are not present in the refractive power measurement. As the Zernike polynomials form an orthogonal function basis, this proves that there are optical aberrations that cannot be captured by the refractive power. Finally, we experimentally demonstrate the validity of our assumption in Sec.~\ref{sec:ShackHartmannMeasurement} by using a Shack-Hartmann sensor to measure the wavefront modulation induced by a high-quality convex lens of well-known refractive power.

\subsection{From Shack-Hartmann Measurements to the Wavefront Aberration Map}
\label{sec:shackHartmann}
\noindent
\begin{figure}[!t]
  \centering
   \includegraphics[width=1.\linewidth]{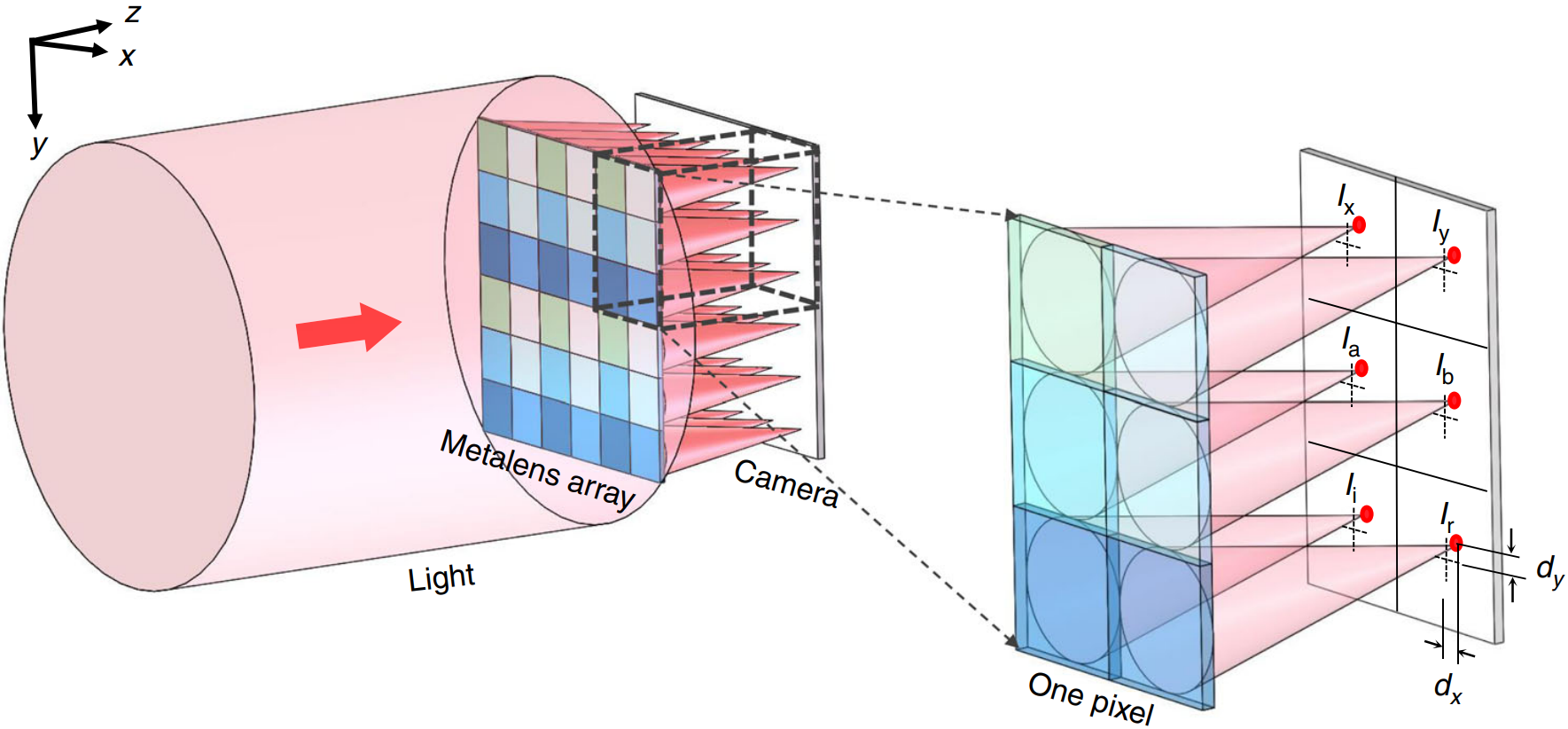}
   \caption{Sketch of the measurement principle of a Shack-Hartmann sensor \cite{Shack_Hartmann}.}
   \label{fig:Shack_Hartmann}
   \vspace{-0.5cm}
\end{figure}
\noindent
Fig.~\ref{fig:Shack_Hartmann} demonstrates the method of operation of a Shack-Hartmann wavefront sensor. If a collimated light beam is transmitting a refractive optical element then the wavefront gets modulated. A Shack-Hartmann sensor consists of a microlens array, which resolves the local wavefront perturbations by focusing a wavefront snippet on a CCD or CMOS sensor. Without any aberrations, the wavefront sensor will capture the light in the center of each pixel. If aberrations are present, then the focusing spot will be displaced locally by $d_{x}$ and $d_{y}$, respectively. The resulting local gradient of the optical path difference ($W$) is given by:
\noindent
\begin{align}
\begin{split}
    \vec{\beta} &\coloneqq \left[\resizebox{0.55\linewidth}{!}{$\begin{array}{cccccc}
         \beta_{x}\Bigr|_{\vec{x}_{1}} &
         \hdots &
         \beta_{x}\Bigr|_{\vec{x}_{m}} &
         \beta_{y}\Bigr|_{\vec{x}_{1}} &
         \hdots &
         \beta_{y}\Bigr|_{\vec{x}_{m}}
    \end{array}$}\right]^{\mathrm{T}}\\[5pt]
    \Leftrightarrow \vec{\beta} &\;= 
    \left[\resizebox{0.775\linewidth}{!}{$
    \begin{array}{cccccc}
         \dfrac{d_{x_{1}}}{\sqrt{f_{\mathrm{sh}}^2 + d_{x_{1}}^2}} &
         \hdots &
         \dfrac{d_{x_{m}}}{\sqrt{f_{\mathrm{sh}}^2 + d_{x_{m}}^2}} &
         \dfrac{d_{y_{1}}}{\sqrt{f_{\mathrm{sh}}^2 + d_{y_{1}}^2}} &
         \hdots &
         \dfrac{d_{y_{m}}}{\sqrt{f_{\mathrm{sh}}^2 + d_{y_{m}}^2}}
    \end{array}$}\right]^{\mathrm{T}}.
\end{split}
\label{eq:OPD_gradients}
\end{align}
\noindent
Here, $f_{\mathrm{sh}}$ denotes the focal length of the microlenses and $m$ specifies the number of microlenses within the array. With the Shack-Hartmann measurement of the local wavefront gradients $\beta_{i}$, the Zernike coefficients $c_{i}$ of the wavefront aberration map are determined by:
\noindent
\begin{align}
\begin{split}
    \vec{\beta} = \dfrac{1}{\rho_{a}} \cdot
    \left[\resizebox{0.35\linewidth}{!}{$\begin{array}{ccc}
         \dfrac{\partial Z_{4}}{\partial \tilde{x}}\Bigr|_{\vec{x}_{1}} & \dots & \dfrac{\partial Z_{n}}{\partial \tilde{x}}\Bigr|_{\vec{x}_{1}}\\
         \vdots & \ddots & \vdots\\
         \dfrac{\partial Z_{4}}{\partial \tilde{x}}\Bigr|_{\vec{x}_{m}} & \dots & \dfrac{\partial Z_{n}}{\partial \tilde{x}}\Bigr|_{\vec{x}_{m}}\\[8pt]
         \dfrac{\partial Z_{4}}{\partial \tilde{y}}\Bigr|_{\vec{x}_{1}} & \dots & \dfrac{\partial Z_{n}}{\partial \tilde{y}}\Bigr|_{\vec{x}_{1}}\\
         \vdots & \ddots & \vdots\\
         \dfrac{\partial Z_{4}}{\partial \tilde{y}}\Bigr|_{\vec{x}_{m}} & \dots & \dfrac{\partial Z_{n}}{\partial \tilde{y}}\Bigr|_{\vec{x}_{m}}
    \end{array}$}\right] \cdot
    \left[\begin{array}{c}
    c_{4}\\
    \vdots\\
    c_{n}
    \end{array}\right] \; \eqqcolon \; \dfrac{1}{\rho_{a}} \cdot \boldsymbol{\mathcal{M}} \cdot \vec{c}\;.
\end{split}
\label{eq:Zernike_numerical_analysis}
\end{align}
\noindent
The Zernike decomposition coefficients $c_{i}$ are uniquely determined if $|\boldsymbol{\mathcal{M}}^{T} \boldsymbol{\mathcal{M}}| \neq 0$. In other words, the Gramian matrix $\boldsymbol{\mathcal{M}}^{T} \boldsymbol{\mathcal{M}}$ has to be invertible, wherefore $\boldsymbol{\mathcal{M}}^{T} \boldsymbol{\mathcal{M}}$ needs to have full rank. If this condition is fulfilled, then the Zernike coefficient vector $\Vec{c}$ can be retrieved from the measured local wavefront gradient vector $\Vec{\beta}$ by:
\begin{equation}
    \Vec{c} \overset{(\ref{eq:Zernike_numerical_analysis})}{=} \rho_{a} \cdot \left[\boldsymbol{\mathcal{M}}^{T} \boldsymbol{\mathcal{M}}\right]^{-1} \cdot \boldsymbol{\mathcal{M}}^{T} \cdot \Vec{\beta}\;.
\end{equation}

\subsection{From Wavefront Aberration Maps to local Refractive Power}
From Sec.~\ref{sec:shackHartmann} we know how to determine the Zernike coefficients $c_{i}$, wherefore we can reconstruct the wavefront aberration map according to Eq.~(\ref{eq:Zernike_decomposition}). If the reference wavefront has been characterized by a plane wave, then the local refractive power of an optical element is given by the second derivative of the wavefront aberration map $W$ with respect to the axis of interest \cite{Wavefront_refractive_power, optical_power_metric}. Hence, the refractive power $D_{x_{i}}$ along the axis $x_{i}$ is given by:
\noindent
\begin{align}
\begin{split}
    D_{x_{i}}(\vec{x}_{a}) &= \dfrac{\partial^2}{\partial x_{i}^2} W(\vec{x}_{a})\;.
\end{split}
\label{eq:refractive_power}
\end{align}
\noindent
Here, the input vector $\vec{x}_{a} \in \mathbb{R}^2$ is restricted to the principal plane of the refractive element. The validity of Equation (\ref{eq:refractive_power}) can be proven for the special case of a spherical thin lens:
\begin{align}
\begin{split}
    f_{x_{a_{1}}}^2 &\;= x_{a_{1}}^2 + \left(f_{x_{a_{1}}} - W(x_{a_{1}})\right)^2 \; \text{, w.l.o.g.:} \;\; x_{a_{2}} \overset{!}{=} 0\\[5pt]
    \Rightarrow W(x_{a_{1}}) &\;= f_{x_{a_{1}}} \resizebox{0.35\linewidth}{!}{$\left( 1 - \sqrt{1 - \left(\dfrac{x_{a_{1}}}{f_{x_{a_{1}}}}\right)^2} \right)$}\\[5pt]
    \Leftrightarrow W(x_{a_{1}}) &\;= f_{x_{a_{1}}} \resizebox{0.69\linewidth}{!}{$\left( 1 - \left(1 - \dfrac{1}{2}\left(\dfrac{x_{a_{1}}}{f_{x_{a_{1}}}}\right)^2 + \mathcal{O}\left\{ \left(\dfrac{x_{a_{1}}}{f_{x_{a_{1}}}}\right)^4 \right\} \right)\right)$}\\[5pt]
    \Rightarrow W(x_{a_{1}}) &\;\approx \dfrac{x_{a_{1}}^2}{2f_{x_{a_{1}}}} \eqqcolon \dfrac{D_{x_{a_{1}}}}{2} \cdot x_{a_{1}}^2\\[5pt]
    \Rightarrow D_{x_{a_{1}}} &\overset{(\ref{eq:refractive_power})}{=} \dfrac{\partial^2}{\partial x_{a_{1}}^2} W(x_{a_{1}}) = D_{x_{a_{1}}}\;. \; \; \; \; \; \; \blacksquare
\end{split}
\label{eq:lemma_proof}
\end{align}
\noindent

\subsection{Information Content of Refractive Power Measurements}
\label{sec:secondDerivative}
Eq.~(\ref{eq:refractive_power}) determines the relationship between refractive power measurements $D$ and wavefront aberration measurements $W$ via the curvature of the optical path difference map. In Sec.~\ref{sec:MTF_intro} we introduced the concept of the PSF as a Fourier optical merit function, which serves as the impulse response function or the Green's function of an optical system \cite{Green_function}. In addition to the Fourier optical approach there is also a ray optics approximation to describe the PSF in terms of the area of a blurring ellipse, which encloses a certain amount of light around the focusing spot in relation to the total amount of energy entering the system through the aperture stop. The area of this blurring ellipse is proportional to the Gaussian curvature \cite{Differential_geometry} of the wavefront aberration map or equivalently speaking, proportional to the determinant of the Hessian matrix of the wavefront aberration function \cite{Wavefront_refractive_power}:
\noindent
\begin{align}
\begin{split}
    \oiint_{\mathcal{C}} \mathrm{PSF}(\Vec{x}_{o}) \Bigr|_{\hat{z}_{o}} \mathrm{d}^2x_{o} \propto \resizebox{0.5\linewidth}{!}{$\left|\left(
    \begin{array}{cc}
        \dfrac{\partial^2}{\partial x_{1}^2} W(\vec{x}_{a}) & \dfrac{\partial}{\partial x_{1}}\dfrac{\partial}{\partial x_{2}} W(\vec{x}_{a}) \\[10pt]
        \dfrac{\partial}{\partial x_{1}}\dfrac{\partial}{\partial x_{2}} W(\vec{x}_{a}) & \dfrac{\partial^2}{\partial x_{2}^2} W(\vec{x}_{a})
    \end{array}\right)\right|\;.$}
\end{split}
\label{eq:Hessian}
\end{align}
\noindent
Here, $\mathcal{C}$ denotes the contour confining the domain of integration, which is given by the blurring ellipse. Due to the relationship presented in Eq.~(\ref{eq:refractive_power}), this matrix is also known as the dioptric power matrix $\boldsymbol{\mathcal{D}}$ \cite{dioptric_power_matrix}. The determinant can be rewritten in terms of the traces of the dioptric power matrix:
\noindent
\begin{align}
\begin{split}
    \oiint_{\mathcal{C}} \mathrm{PSF}(\Vec{x}_{o}) \Bigr|_{\hat{z}_{o}} \mathrm{d}^2x_{o} \overset{(\ref{eq:Hessian})}{\propto} \dfrac{1}{2} \left[ \left(\mathrm{tr}\left(\boldsymbol{\mathcal{D}}\right)\right)^2 - \mathrm{tr}\left(\boldsymbol{\mathcal{D}}^2\right) \right]\;.
\end{split}
\label{eq:Hessian_Traces}
\end{align}
\noindent
So far, the automotive industry exclusively specifies requirements in terms of the refractive power w.r.t. the horizontal and vertical directions. Consequently, only the trace of $\boldsymbol{\mathcal{D}}$ is measured and off-diagonal elements in the Hessian matrix are not investigated. This demonstrates that there is a blind spot in the quality assurance chain at the moment.
This conclusion can be further underpinned by an mathematical argument. The trace of $\boldsymbol{\mathcal{D}}$ is given by:
\noindent
\begin{align}
\begin{split}
    \mathrm{tr}\left(\boldsymbol{\mathcal{D}}\right) = \sum \limits_{i=1}^{d} D_{x_{i}}(\vec{x}_{a}) = \Laplace W(\vec{x}_{a})\;.
\end{split}
\label{eq:Laplace}
\end{align}
\noindent
Consequently, the trace of $\boldsymbol{\mathcal{D}}$ is unaffected by wavefront aberration fields, which fulfill the Laplace equation:
\noindent
\begin{equation}
    \Laplace \Gamma(\vec{x}_{a}) \overset{!}{=} 0\;.
    \label{eq:Laplace_equation}
\end{equation}
\noindent
As a result, the trace of $\boldsymbol{\mathcal{D}}$ is Gauge invariant under aberration fields $\Gamma(\vec{x}_{a})$ that are composed of harmonic functions. Hence, Zernike polynomials in Table~\ref{tab:Zernike_polynomials} that are harmonic functions (like astigmatism or trefoil) will not alter the trace of $\boldsymbol{\mathcal{D}}$.

In a nutshell, refractive power measurements are not sensitive for optical distortions quantified by $c_{1}$ and $c_{2}$. Furthermore, the refractive power is invariant under oblique astigmatism given by $c_{3}$ if the refractive power requirements are specified exclusively along the horizontal and vertical axis, as it is the current governing standard in the automotive industry. Finally, those quality standards are insufficient for extracting more fundamental information about the optical system in terms of the $\mathrm{PSF}$. Nonetheless, the aberrations associated with these polynomials have been proven to have an influence on the performance of AI algorithms \cite{Mueller_dist_shift, Mueller_sim}.

\subsection{Experimental Verification}
\label{sec:ShackHartmannMeasurement}
\begin{figure}[!t]
  \centering
   \includegraphics[width=1.\linewidth]{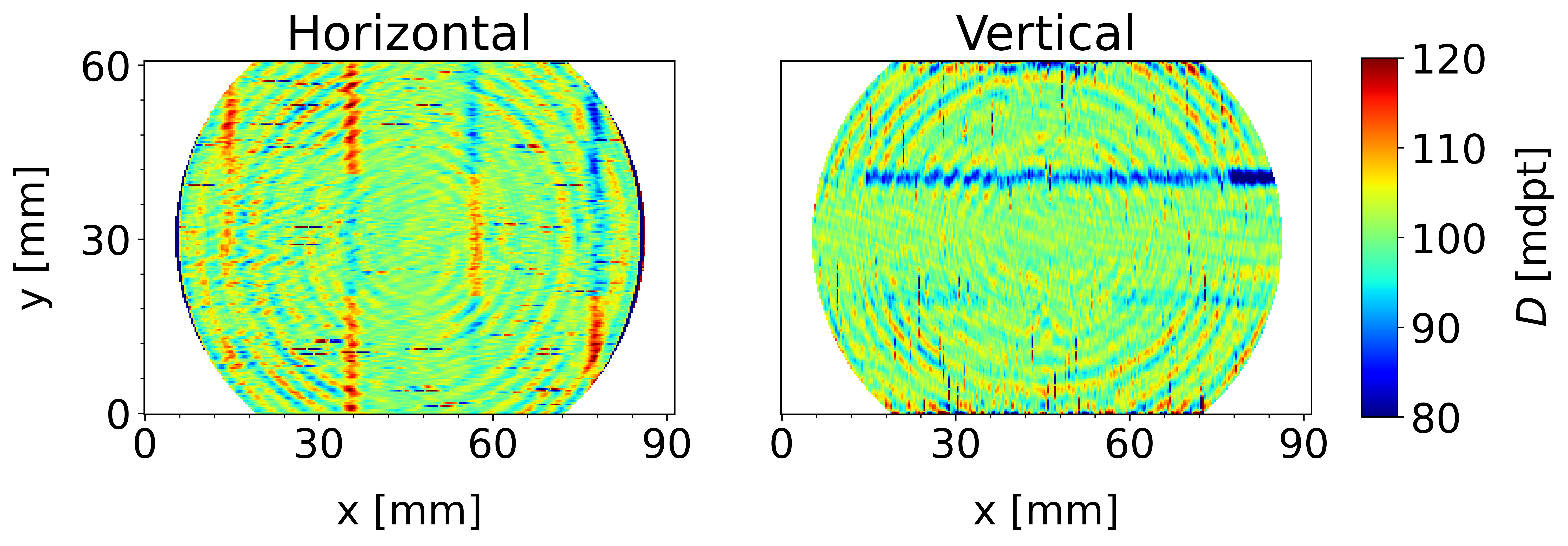}
   \caption{Wavefront measurement performed on a ${\left< D \right>=100.3}$\Unit{mdpt} reference lens. In order to cover the entire aperture of the lens, several Shack-Hartmann measurements have been stitched together. This procedure has introduced artifacts, which are visible in the measurement data by strongly pronounced vertical and horizontal lines. In total, 15 measurements have been performed over the calibration lens aperture of $d=10$\Unit{cm}.}
   \label{fig:Wavefront_measurment}
   \vspace{-0.5cm}
\end{figure}

Since Eq.~(\ref{eq:refractive_power}) is not well established in the automotive industry, we experimentally demonstrate the validity of the relationship by a Shack-Hartmann wavefront measurement of a calibration lens. The lens under test was produced by Zeiss and is traced back to national standards by an accredited calibration authority. The local wavefront gradients $\beta_{i}$ are measured by a Shack-Hartmann sensor and the refractive power is retrieved by utilizing Eq.~(\ref{eq:refractive_power}). As demonstrated by Eq.~\ref{eq:Zernike_numerical_analysis} the Shack-Hartmanm measurement yields the first derivative of the wavefront. Here, we measure the lens and numerically determine the second derivative by a simple central difference scheme, which should result in the specified refractive power. Fig.~\ref{fig:Wavefront_measurment} illustrates the outcome w.r.t. the refractive power map over the lens aperture. From the frequency distribution of the local refractive power across the entire principal plane, the expectation value for the global refractive power of the optical element in the $x$- and $y$-plane can be deduced. The expectation values meet the certified refractive power values of the calibration lens within the uncertainty intervals. Hence, the validity of Eq.~(\ref{eq:refractive_power}) has also been experimentally confirmed.
\noindent
\noindent

Summarizing, we have demonstrated that fundamentally several optical aberrations are not captured by a refractive power measurement. The image quality can be deteriorated even though the refractive power measurement indicates a compliant windscreen sample. From previous studies on the effect of oblique astigmatism ($c_{3}$) on road sign classification \cite{Mueller_dist_shift, Mueller_sim} it becomes evident, that refractive power measurements are insufficient for specifying the quality of a windshield in order to ensure reliable computer vision for autonomous driving vehicles.

\section{Modulation Transfer Function}
\label{sec:MTF}
In this section, we will demonstrate why the windscreen and the camera form a joint optical system that cannot be separated into two independent constituents, such that the MTF cannot be determined for the two systems separately. First, we argue how the refractive power of the windscreen interacts with the focal length of the camera system. In a second step, this is experimentally verified using a MTF measurement with and without a windscreen. A discussion elaborates on several implications for the production and testing process.

\subsection{Field Curvature}
\begin{figure}[!b]
    \centering
    \includegraphics[width=0.85\linewidth]{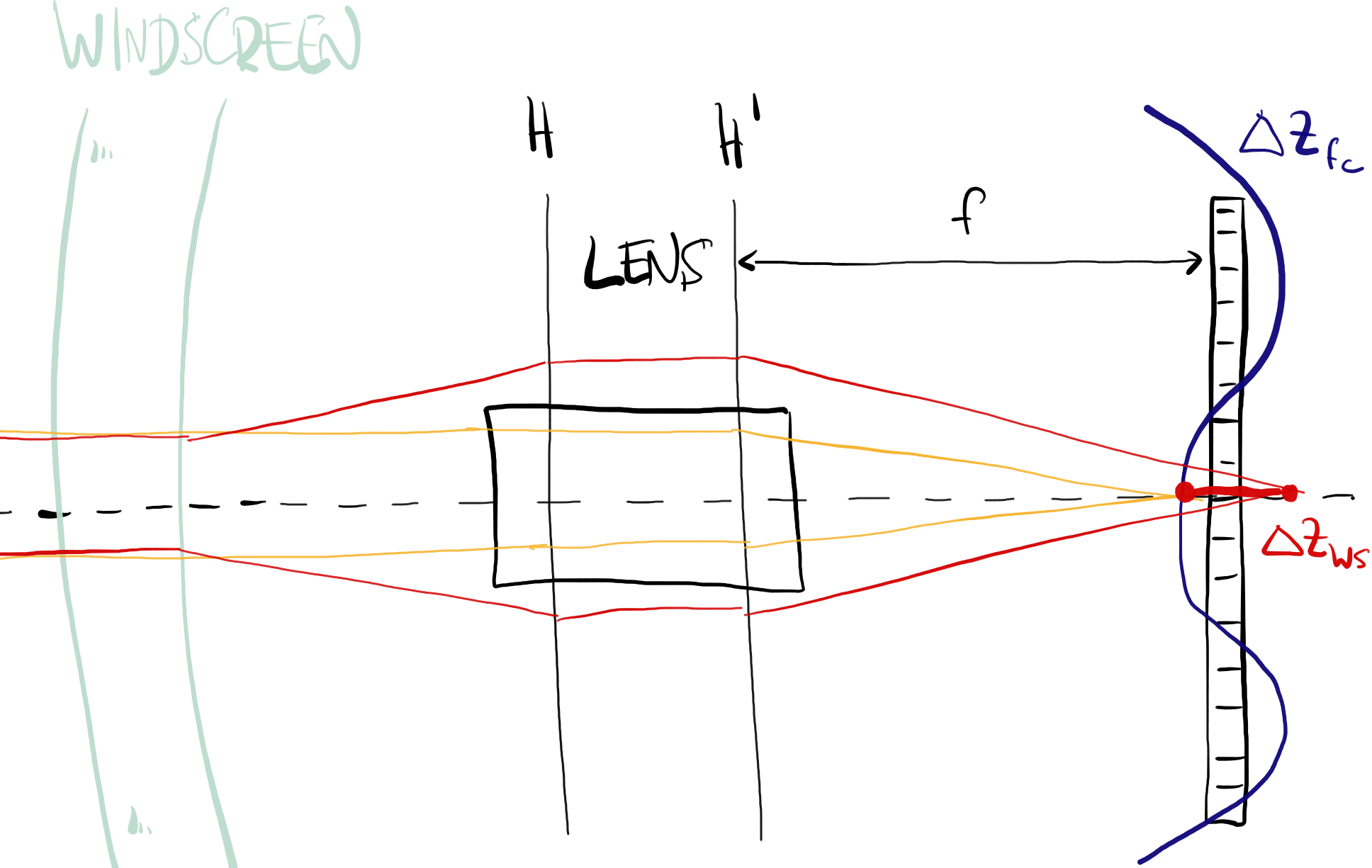}
    \caption{Windscreen and lens form a joint optical system. $H$ and $H^\prime$ are the principle planes of the lens, $f$ is the nominal focal length. The blue line visualizes the field curvature (not to scale). Normally, parallel rays are focused onto the field curvature (yellow line). Windscreen refractive power shortens or prolongs the effective focal length of the lens (red line). There are two different focus offsets $\Delta z_\text{fc}$ and $\Delta z_\text{ws}$ which may add or even cancel at different fields of view.}
    \label{fig:WSS_fieldCurvature}
    \vspace{-0.125cm}
\end{figure}

The focal length of an imaging system varies over the field of view, with the so-called \emph{field curvature} being a prominent optimization goal for any lens designer. The semiconductor production processes produce completely flat image sensors, which for the imaging optics is a challenge, as the field curvature needs to be flat as well to minimize aberrations. This field curvature, as a design property of the lens, is given by the offset $\Delta z_\text{fc}$ over field in units of length, typically on the micrometer range. A symbolic field curvature is visualized in Fig.~\ref{fig:WSS_fieldCurvature}.
\noindent
\noindent
As explained above, the refractive power of the windscreen leads to parallel rays converging or diverging. Taking the two elements windscreen and lens together yields a second focus offset $\Delta z_\text{ws}$ for the camera system, as the converging (diverging) rays will shorten (prolong) the focal length of the camera system. Fig.~\ref{fig:WSS_fieldCurvature} depicts this situation. The two offsets are added for the system offset, such that:
\begin{equation}
    \Delta z = \Delta z_\text{ws} + \Delta z_\text{fc}\;.
\end{equation}
Importantly, both $\Delta z_\text{ws}$ and $\Delta z_\text{fc}$ can have positive or negative values, and thus the overall offset may vanish when these terms cancel. A vanishing offset value implies a sharpening of the system. Here, a MTF measurement of the camera alone would yield a certain number, while putting a windscreen in front of the camera would act like glasses and the image would become sharper. That this is indeed the case in practice is presented in the following section.


\subsection{Experimental Validation}
\begin{figure}[!t]
\begin{subfigure}{0.49\linewidth}
  \centering
  \includegraphics[width=1.\linewidth]{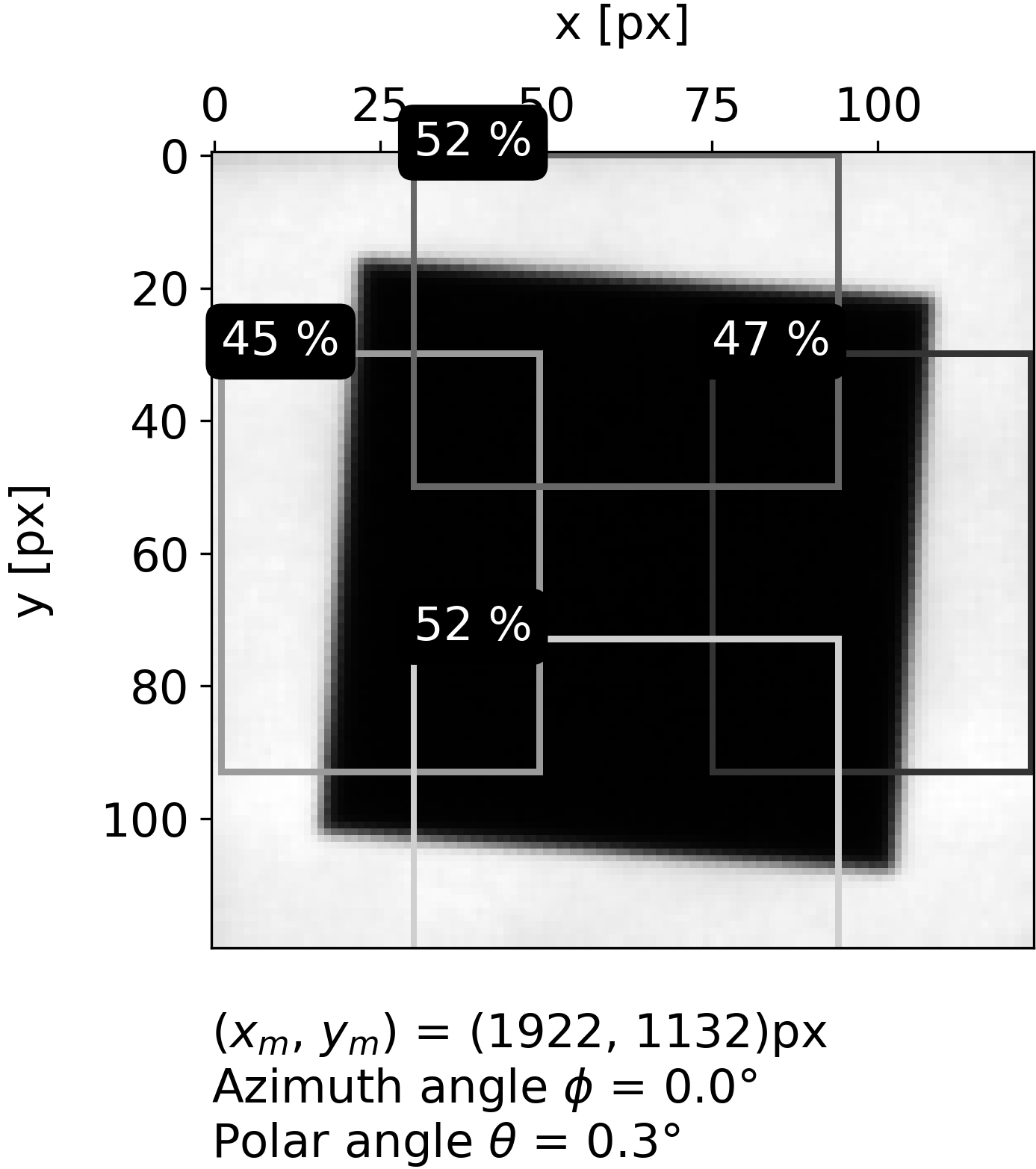}
  \caption{ADAS camera}
  \label{fig:MTF_camera_only}
\end{subfigure}
\hfill
\begin{subfigure}{0.49\linewidth}
  \centering
  \includegraphics[width=1.\linewidth]{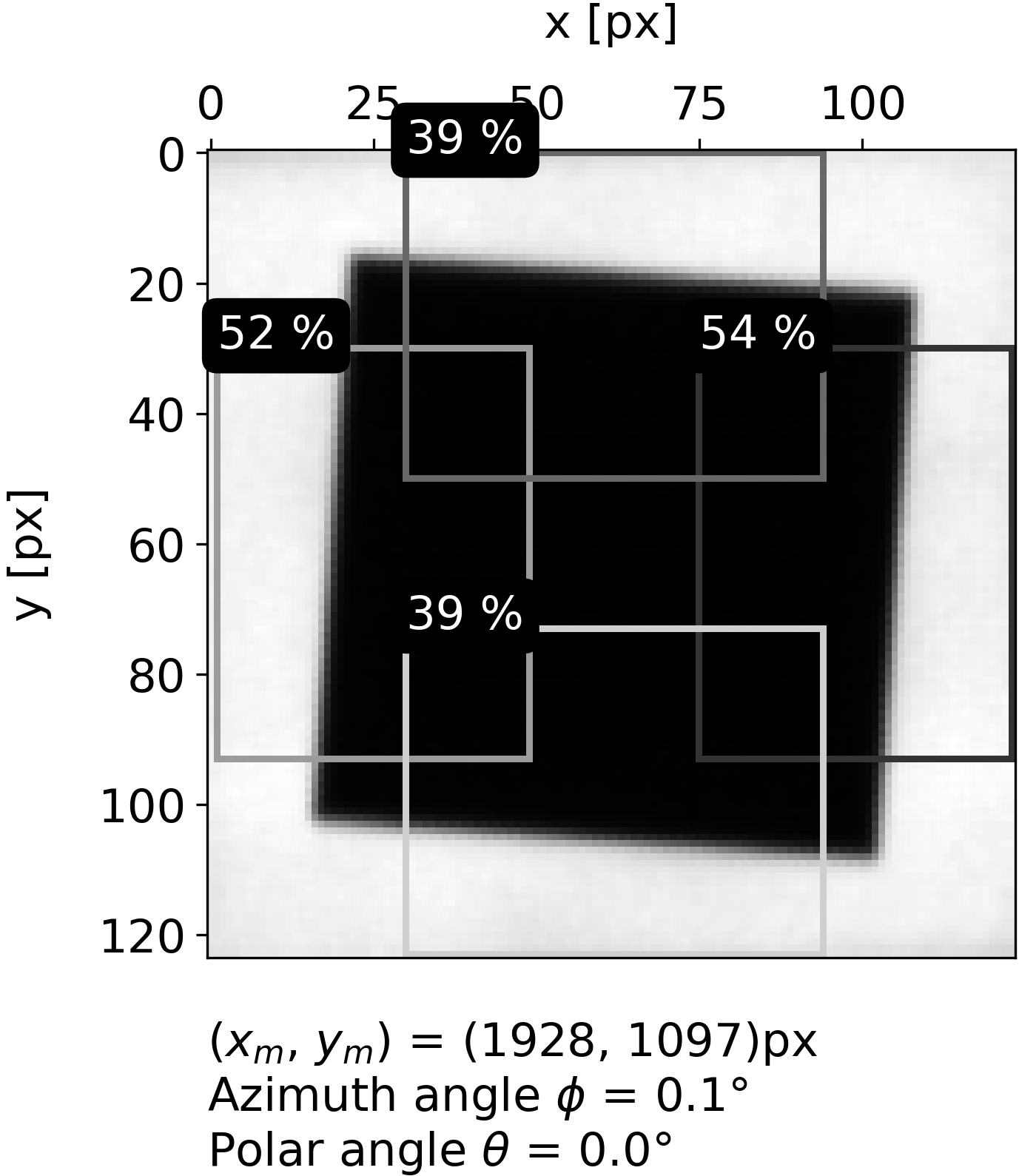}
  \caption{ADAS camera + windshield}
  \label{fig:MTF_windscreen}
\end{subfigure}
\caption{MTF measurement for an ADAS system based on the Slanted Edge method according to ISO 12233 \cite{ISO12233}.}
\label{fig:MTF_slanted_edge}
\vspace{-0.5cm}
\end{figure}
Fig.~\ref{fig:MTF_slanted_edge} depicts two slanted edge measurements, one without a windscreen (\ref{fig:MTF_camera_only}), and one with a windscreen placed in front of the camera system (\ref{fig:MTF_windscreen}). The insets indicate the MTF values derived from the numerical evaluation for all four edges, using an ISO12233-compliant algorithm \cite{ISO12233}. There are two horizontal and two vertical values. The two vertical values (top and bottom) distinctly decrease from $52\pm1.5\Unit{\%}~[95\Unit{\%}]$ to $39\pm1.7\Unit{\%}~[95\Unit{\%}]$ when a windscreen is placed in front of the camera. However, for the horizontal direction (left and right) the MTF values both significantly increase from $45/47\pm1.5\Unit{\%}~[95\Unit{\%}]$ to $52/54\pm1.6\Unit{\%}~[95\Unit{\%}]$ when the windscreen is placed in front of the camera. The results experimentally confirm that the defocus $\Delta z_\text{ws}$ and $\Delta z_\text{fc}$ may cancel to a certain degree, increasing the sharpness like glasses would do for a myopic person. This conclusion is well established in physics \cite{Eugene_Hecht} for decades but the implications for the quality assurance testing procedure of ADAS systems in the automotive industry are not well prevalent.

\subsection{Discussion}
Both the field curvature of the lens and the refractive power of the windscreen are spatially variant. The field curvature not only varies over field (radially) as the name implies, but due to production tolerances the rotation symmetry of the lens is usually broken to a certain degree.
The field of view of the lens projected on the windscreen yields a trapezoidal cutout (cf. Fig.~\ref{fig:optical_power_measurement}). I.e., the (almost) rotational symmetry of the lens projected on the windscreen combines with the local refractive power variation of the windscreen in this cutout. Not only that, but the windscreen also has distinctly different refractive power for the horizontal and the vertical direction as given by the Kerkhof model \cite{Metrologia}.

Taken together, it is apparent that a windscreen cannot be qualified by a MTF measurement if both the windscreen and the camera are measured separately. The experimental sharpening unambiguously demonstrates a non-linear process, proving that the two elements cannot be separated using linear system theory (read: the individual MTFs cannot be multiplied). The way individual production tolerances will add is not predictable. In brief, it is not possible to determine individual MTF limits, as the combination of individual tolerances may hold both good and bad surprises, either sending a good system to scrap or a bad system to the field.

Therefore, any solution using MTF would have to measure the MTF on the combined system of produced windscreen and camera system with their individual production tolerances. This could be either at the production site of the Tier~1 or the OEM. But there are still several important open questions that make this an unattractive proposal: if an assembly is non-compliant, is it worth finding a compliant combination, does it make economically sense? How big are the assembly tolerances fitting the windscreen into the car body? If the OEM wants the measurement system at the site of the Tier~1, one should be aware that the assembly of the windscreen into the car produces distinct mechanical tolerances, changing the shape and internal tension of the windscreen. As we are looking for subtle differences in optical quality, this may affect the pre-assembled camera system as well. Finally, from an automotive process view it is clear that an independent measurement of the camera and the windscreen is much preferred.

Summarizing, the MTF is a measure of 'sharpness' based on linear system theory. The windscreen and the camera form a combined optical system that cannot be separated, which prohibits its use for windscreen characterization without the actual, produced camera system in place. Taken together with the possibility of finding a better metric we are skeptical that the MTF should be prioritized for windscreen characterization going forward.

\section{Simulating Optical Properties}
Having shown that basically no current measurement system in the automotive windscreen industry is capable of a meaningful characterization of the windscreen optical quality for downstream AI algorithm consumption, what could be a way forward? A comprehensive experimental study using thousands of actual cameras and windscreens is out of the question. Therefore, the AI performance needs to be linked to the windscreen optical quality by simulation, using physical-realistic optic models. These simulations need to model the production tolerances of both windscreens and cameras. Then, the performance requirements of the AI-based ADAS functionalities can be translated to optical quality specifications for windscreen production.

The waveform description is fundamental and includes all optical effects and aberrations, and can be measured by a Shack-Hartmann sensor. Currently, this is not a viable approach to windscreen characterization at the site of the Tier~1, as it is too expensive and more importantly, too slow for a 100\Unit{\%} part check. Nonetheless, we believe it is possible to use special laboratory-grade equipment to create the physical-realistic optical models necessary for the simulations, and then derive from these simulations an understanding of the optical properties that are really necessary for the AI performance. Finally, from this we can deduce a simplified form of measurement that captures this newfound knowledge of the required optical properties. A first example of this process is published in \cite{krebs}.

Therefore, the challenge is to understand those optical properties that are really necessary for a robust AI algorithm performance. We believe that this is a necessary step, and without it, the move from ADAS to AD will be prohibitively difficult, as production tolerances combined with the complexity of the world create an unmanageable number of combinations.


\section{Summary}
In automotive mass production, the inspection systems at the suppliers need to measure the quality of windscreens in a meaningful way for the final device performance. Modern ADAS and future AD camera systems are based on AI algorithms, wherefore the windscreen quality needs to be linked to the performance of these algorithms. Currently, there are two types of measurements established in the industry to measure the optical quality of a windscreen: refractive power and MTF. In this article, we demonstrated how both these measurements are fundamentally not capable of capturing relevant optical properties of the windscreen.

The refractive power measurement does not include several aberrations -- given by Zernike polynomials, e.g. oblique astigmatism -- while these aberrations obviously affect the performance of the AI algorithms: oblique astigmatism causes a directional blurring of the scene, and blurring causes a degradation of the performance. Because of the orthogonality of the Zernike polynomials, it is clear that this information is simply lacking in refractive power measurements.

MTF is based on linear system theory, where independent optical systems might be multiplied in frequency space to yield the system MTF. This is, for example, the case for the lens and the imager. Here, we demonstrated mathematically and experimentally that the windscreen forms a novel optical system together with the lens of the camera system, which cannot be separated into individual components. Therefore, measuring the MTF on the windscreen alone will not yield the performance of the combined system. Thus, the final assembly of the windscreen and camera system in the car may be both better or worse than the EOL measurement at the windscreen production site, either sending good parts to scrap or bad parts into the field.

Every car has a windscreen. Using the knowledge presented in this article we believe that the automotive industry needs to focus their efforts on finding novel measurement methods that qualify the optical quality of windscreens in a meaningful way for the downstream AI algorithms. We propose a concept using fundamental wave (and Fourier) optics to characterize the windscreens and combine wavefront measurements and physical-realistic simulations to reach an understanding of what optical properties are really important for AI computer vision algorithms.
We believe that it cannot be said generally that the optical quality of windscreens is too low
-- what is currently lacking is not optical quality, but understanding about the robustness of the algorithms against optical aberrations. We simply do not know what optical quality is needed exactly. Taking these elements together we believe that a novel metric can be found that contains the relevant information, while at the same time the measurement is practical enough to be used stand-alone at the windscreen production site. This is the great windscreen challenge the automotive industry currently faces.


\bibliographystyle{ieee_fullname.bst}
\bibliography{windscreen_optical_quality}

\begin{thebibliography}{10}\itemsep=-1pt

\bibitem{MTF_aliasing}
Uwe Artmann.
\newblock Quantify aliasing a new approach to make resolution measurement more
  robust.
\newblock {\em Electronic Imaging}, 31(10):320, 2019.

\bibitem{Boreman}
Glenn~D. Boreman.
\newblock {\em Modulation Transfer Function in Optical and Electro-Optical
  Systems, Second Edition}.
\newblock SPIE Press, 2001.

\bibitem{born_wolf}
Max Born, et~al.
\newblock {\em Principles of Optics: Electromagnetic Theory of Propagation,
  Interference and Diffraction of Light}.
\newblock Cambridge University Press, 7 edition, 1999.

\bibitem{Braun_tm_2022}
Alexander Braun.
\newblock Automotive mass production of camera systems: Linking image quality
  to {AI} performance.
\newblock {\em tm - Technisches Messen}, 2022.

\bibitem{Lumetrics}
David Compertore.
\newblock Adas windshield measurements - white paper.
\newblock Technical report, Lumetrics, 2021.

\bibitem{DIN52305}
DIN:52305.
\newblock Determination of the optical deviation and refractive power of safety
  glass for vehicle glazing.
\newblock Technical report, German Institute for Standardization, 1995.

\bibitem{Windscreens_and_safety_III}
DIN:52353.
\newblock Pr{\"u}fung der abriebfestigkeit von fahrzeugverglasung mit dem
  wischer-test.
\newblock Technical report, Deutsches Institut f{\"u}r Normung (DIN), 2019.

\bibitem{Differential_geometry}
Manfredo~Perdig{\~a}o do Carmo.
\newblock {\em Differential geometry of curves and surfaces}.
\newblock Prentice-Hall, 1976.

\bibitem{ECE43}
ECE:43.
\newblock Uniform provisions concerning the approval of safety glazing
  materials and their installation on vehicles.
\newblock Technical report, Regulation No. 43 of the Economic Commission for
  Europe of the United Nations (UN/ECE), 2014.

\bibitem{Goodman}
Joseph~W. Goodman.
\newblock {\em Introduction to Fourier optics}.
\newblock Stanford University / McGraw-Hill, 1968.

\bibitem{dioptric_power_matrix}
William Harris.
\newblock The matrix representation of dioptric power. part 1: An introduction.
\newblock {\em South African Optometrist}, 47:19--23, 01 1988.

\bibitem{Eugene_Hecht}
Eugene Hecht.
\newblock {\em Optik}.
\newblock De Gruyter, 2018.

\bibitem{heizmann:22}
Michael Heizmann, et~al.
\newblock Implementing machine learning: chances and challenges.
\newblock {\em at -- {A}utomatisierungstechnik}, 70(1):90--101, 2022.

\bibitem{IEEEP2020}
IEEE:P2020.
\newblock Automotive imaging - white paper.
\newblock Technical report, IEEE Standards Association, 2018.

\bibitem{Windscreens_and_safety_II}
M.~J. Irland.
\newblock Windshield optics.
\newblock {\em Applied Optics}, 8(9):1787--1790, 1969.

\bibitem{ISO12233}
ISO:12233.
\newblock Photography - electronic still picture imaging - resolution and
  spatial frequency responses.
\newblock Technical report, International Organization for Standardization,
  2023.

\bibitem{ISO24157}
ISO:24157.
\newblock Ophthalmic optics and instruments - reporting aberrations of the
  human eye.
\newblock Technical report, International Organization for Standardization,
  2008.

\bibitem{ISO16949}
ISO/TS:16949.
\newblock Quality management systems --- particular requirements for the
  application of iso 9001:2008 for automotive production and relevant service
  part organizations.
\newblock Technical report, International Organization for Standardization,
  2009.

\bibitem{MTF_wedges}
Norman Koren, et~al.
\newblock Measuring mtf with wedges: pitfalls and best practices.
\newblock {\em Electronic Imaging}, 29(19):6, 2017.

\bibitem{krebs}
Christian Krebs, et~al.
\newblock Impact of windshield optical aberrations on visual range camera based
  classification tasks performed by cnns.
\newblock {\em Proc. London Imaging Meeting: Imaging for Deep Learning}, pages
  83--87, 2021.

\bibitem{Green_function}
Prem~K. Kythe.
\newblock {\em Green's Functions and Linear Differential Equations: Theory,
  Applications, and Computation}.
\newblock CRC Press, 2011.

\bibitem{LaVision}
LaVision.
\newblock Automotive imaging for safer driving.
\newblock Technical report, LaVision, 2023.

\bibitem{ISRA}
Thomas Mitra.
\newblock Benefits of optical distortion measurement.
\newblock Technical report, ISRA Vision, 2022.

\bibitem{Mueller_sim}
Patrick M{\"u}ller et~al.
\newblock Simulating optical properties to access novel metrological parameter
  ranges and the impact of different model approximations.
\newblock In {\em 2022 IEEE International Workshop on Metrology for Automotive
  (MetroAutomotive)}, pages 133--138, 2022.

\bibitem{Braun_MTF_performance}
Patrick M{\"u}ller et~al.
\newblock Mtf as a performance indicator for ai algorithms?
\newblock {\em Electronic Imaging}, 35(AVM-125), 2023.

\bibitem{Mueller_dist_shift}
Patrick M{\"u}ller, et~al.
\newblock Impact of realistic properties of the point spread function on
  classification tasks to reveal a possible distribution shift.
\newblock {\em NeurIPS Workshop DistShift}, 2022.

\bibitem{MTF_stability_repeatability}
J. Roland.
\newblock A study of slanted-edge mtf stability and repeatability.
\newblock In {\em Electronic imaging}, 2015.

\bibitem{steger:18a}
Carsten Steger, et~al.
\newblock {\em Machine Vision Algorithms and Applications}.
\newblock Wiley-VCH, Weinheim, 2nd edition, 2018.

\bibitem{Wavefront_refractive_power}
Larry~N. Thibos.
\newblock Calculation of the geometrical point-spread function from wavefront
  aberrations.
\newblock {\em Ophthalmic and Physiological Optics}, 39(4):232--244, June 2019.

\bibitem{optical_power_metric}
L.~N. Thibos, et~al.
\newblock Accuracy and precision of objective refraction from wavefront
  aberrations.
\newblock {\em Journal of Vision}, 4(4):9, April 2004.

\bibitem{VDA6.3}
VDA:6.3.
\newblock Qualit{\"a}tsmanagement in der automobilindustrie - prozessaudit.
\newblock Technical Report~4, Verband der Deutschen Automobilindustrie (VDA),
  2023.

\bibitem{BMW}
Korbinian Weik, et~al.
\newblock Imaging through curved glass: windshield optical impact on automotive
  cameras.
\newblock {\em Proceedings of SPIE}, 12231(122310A), 2022.

\bibitem{Metrologia}
Dominik~W. Wolf, et~al.
\newblock Optical power measurement in the automotive world.
\newblock {\em Currently under review at Metrologia}, 2023.

\bibitem{Shack_Hartmann}
Z. Yang, et~al.
\newblock Generalized hartmann-shack array of dielectric metalens sub-arrays
  for polarimetric beam profiling.
\newblock {\em Nature Commun}, 9(4607), 2018.

\end{thebibliography}
\end{document}